# Interactive Diabetes Risk Prediction Using Explainable Machine Learning: A Dash-Based Approach with SHAP, LIME, and Comorbidity Insights

Udaya Allani, Department of Computer Science, University of Arkansas at Little Rock, Arkansas, USA

*Abstract*— **This paper presents an interactive machine learning-based system for diabetes risk prediction and health assessment using the Behavioral Risk Factor Surveillance System (BRFSS) dataset. The proposed framework integrates data preprocessing, feature engineering, class imbalance handling, and explainable AI to generate both predictive and personalized health insights. Multiple classifiers, including Logistic Regression, Random Forest, XGBoost, and LightGBM, were evaluated across three sampling strategies: original, SMOTE oversampling, and random undersampling. A LightGBM model trained on an undersampled dataset was selected based on recall-optimized cross-validation performance. To enhance interpretability, SHAP and LIME were applied to explain both global and individual predictions. The system also introduces composite lifestyle and healthcare scores and derives comorbidity risk insights using correlation analysis. The final model is deployed as a multi-step Dash web application supporting interactive inputs, risk visualization, and personalized recommendations. This work demonstrates the practical potential of explainable AI to support public health screening and awareness, especially in underserved populations.**

## I. INTRODUCTION

Diabetes is a growing public health concern, with significant implications for long-term health outcomes and healthcare access. Early identification of individuals at risk can enable timely lifestyle interventions, reduce complications, and support population-level disease prevention. The Behavioral Risk Factor Surveillance System (BRFSS) provides a large-scale, self-reported dataset capturing health-related behaviors, chronic conditions, and access to care making it a valuable resource for data-driven health analytics.

Machine learning (ML) techniques have shown promise in analyzing such complex health datasets to predict chronic disease risks, including diabetes. However, the lack of model transparency and interpretability has limited their adoption in real-world healthcare settings. To address this, explainable artificial intelligence (XAI) approaches such as SHAP

(SHapley Additive exPlanations) and LIME (Local Interpretable Model-Agnostic Explanations) have been introduced to demystify model outputs and improve trust among stakeholders.

This paper presents an end-to-end pipeline for diabetes risk prediction using the BRFSS dataset. The proposed framework includes: (1) a structured preprocessing pipeline with feature engineering and class imbalance handling; (2) evaluation of multiple machine learning models across SMOTE and undersampled datasets, with a recall-optimized LightGBM model trained on the undersampled data selected for final deployment; (3) integration of SHAP and LIME for dual-level interpretability; and (4) a deployed interactive Dash application for real-time risk analysis. The system also introduces composite lifestyle and healthcare scores to aid understanding and identifies comorbidity risks through correlation-based insights.

In contrast to existing studies focused primarily on accuracy metrics, this work emphasizes transparency, interpretability, and actionable outputs through an accessible web interface. It is intended to support both individuals and public health efforts in understanding diabetes risk factors and promoting preventive care.

## II. RELATED WORK

Numerous studies have leveraged the Behavioral Risk Factor Surveillance System (BRFSS) dataset to develop machine learning (ML) models for diabetes prediction. These efforts highlight the potential of data-driven approaches in public health, but vary significantly in methodology, feature selection, model interpretability, and deployment.

Chowdhury et al. [1] applied various sampling techniques such as SMOTE, SMOTE-Tomek, and SMOTE-EN alongside models like Logistic Regression, AdaBoost, and Gradient Boost to address class imbalance in BRFSS 2021 data. While the study emphasized recall improvement, it lacked explainability components, making it difficult to understand model decisions.

Liu et al. [2] conducted a comparative analysis of Logistic Regression, Random Forest, and XGBoost using BRFSS 2015 data. They employed SMOTE for balancing and used SHAP to identify key predictors like general health and blood pressure. However, their study did not explore LIME, nor did it integrate statistical evaluation across multiple sampling methods.

Ahmed et al. [3] proposed an explainable ML framework using Logistic Regression and Random Forest, integrating SHAP and LIME to provide local and global model explanations. Although they achieved 86% accuracy and presented compelling visual interpretations, they did not address the class imbalance problem in the dataset. Furthermore, their framework lacked real-time application, composite scoring, and comorbidity risk analysis.

Nguyen and Zhang [4] utilized Decision Tree, K-Nearest Neighbors, and Logistic Regression on BRFSS data to build interpretable models. Their focus remained on basic performance metrics without using advanced sampling techniques or XAI methods.

In contrast to these studies, the present work integrates advanced preprocessing, undersampling, and a LightGBM classifier selected through cross-validation based on recall. Additionally, it combines SHAP and LIME for both global and local interpretability, introduces lifestyle and healthcare scores for personalized risk feedback, and deploys the model in a web-based Dash application. A further contribution includes correlation-based comorbidity insights, which have not been addressed in prior BRFSS-based research.

## III. METHODOLOGY

### A. Dataset Description

This study uses a subset of the 2015 Behavioral Risk Factor Surveillance System (BRFSS) dataset, a health-related telephone survey conducted by the Centers for Disease Control and Prevention (CDC). After preprocessing and feature selection, the working dataset included approximately 250,000 self-reported responses and 22 relevant features covering demographic, lifestyle, medical, and healthcare-related variables. The original target variable in the dataset included three classes: diabetic, non-diabetic, and prediabetic. These features were selected to support diabetes risk prediction using machine learning techniques.

### B. Data Preprocessing

The BRFSS dataset underwent a structured preprocessing pipeline to ensure data quality and consistency prior to modeling. No missing values were present in the dataset, and although some extreme BMI values were observed, they were retained to reflect realistic variation in self-reported health metrics. Duplicate records were identified and removed to reduce redundancy. Categorical variables such as income were encoded using ordinal encoding, and all columns were converted to appropriate data types to ensure compatibility with machine learning frameworks.

The original target variable, Diabetes_012, included three classes: non-diabetic (0), prediabetic (1), and diabetic (2). To frame the task as binary classification, prediabetic entries were excluded. The remaining diabetic (2) entries were re-coded as 1, and non-diabetic (0) entries remained unchanged. This yielded a class distribution of approximately 84.41% non-diabetic and 15.59% diabetic. The resulting dataset was clean, structured, and suitable for training and evaluation of machine learning models.

### C. Feature Engineering

To improve model performance and interpretability, several engineered features were introduced:

- **Lifestyle Score**: A composite score combining behavioral indicators such as physical activity, fruit and vegetable intake, smoking, alcohol consumption, and days of poor mental and physical health. The score was normalized and scaled from 1 to 5.

- **Healthcare Access Score**: Aggregated from responses related to insurance coverage, ability to afford care, and general access to healthcare services.

- **Risk Factor Count**: Created by summing binary indicators of high-risk conditions: HighBP, HighChol, Stroke, and HeartDiseaseorAttack. This count provided a cumulative measure of an individual's chronic health burden and emerged as the most influential predictor in SHAP analysis.

These derived features improved the model's ability to capture health behavior patterns and clinical risk profiles beyond individual variables.

### D. Handling Class Imbalance

Due to the natural imbalance between diabetic and non-diabetic cases, three sampling strategies were employed for comparison:

1. **Original dataset** without any sampling adjustment.
2. **SMOTE (Synthetic Minority Oversampling Technique)** to synthetically balance classes by generating new minority samples.
3. **Random Undersampling**, which reduced the majority class to match the minority class.

Each approach was applied independently, and model performance was evaluated using cross-validation to determine the most effective technique. Ultimately, undersampling combined with LightGBM yielded the highest recall and was selected as the final strategy.

### E. Model Training and Evaluation

Several machine learning models were tested, including Logistic Regression, Decision Tree, Random Forest, XGBoost, and LightGBM. To address class imbalance, each model was evaluated across three sampling strategies: original distribution, SMOTE oversampling, and random undersampling. The evaluation focused on recall as the primary metric, aiming to minimize false negatives in a healthcare setting.

Hyperparameter tuning was performed using GridSearchCV to optimize each model's performance within each sampling strategy. To assess the statistical significance of performance differences, a one-way ANOVA was conducted on recall scores obtained via stratified k-fold cross-validation, followed by Tukey's HSD post-hoc test to identify specific model differences.

LightGBM combined with random undersampling demonstrated the highest and most consistent recall, with statistically significant improvements over other models. Due to its balance of performance and explainability, it was selected as the final deployed model.

## F. Explainable AI Integration

To enhance transparency and support decision-making, both SHAP (SHapley Additive exPlanations) and LIME (Local Interpretable Model-Agnostic Explanations) were integrated. SHAP was used for global feature importance visualization and interpretation, revealing that the Risk Factor Count, General Health, and BMI were among the top predictors. LIME was used to explain individual predictions by approximating the local behavior of the model around specific instances. Together, these tools provided a comprehensive view of how predictions were generated and why.

## G. Comorbidity Correlation Analysis

To extend the system's utility, a correlation-based analysis was conducted to identify comorbidities most associated with diabetes. Pearson correlation coefficients were calculated between diabetes and related variables such as high blood pressure, high cholesterol, stroke, and heart disease. These insights were integrated into the application's output as risk flags and visual summaries to support further health risk assessment.

## H. Dash Application Deployment

The entire pipeline was deployed as a multi-step interactive web application using Dash (Plotly). The user interface guides individuals through sequential stages basic information, lifestyle habits, healthcare access, and medical history. Upon submission, the system generates real-time predictions, personalized health recommendations, comorbidity insights and alerts, and LIME/SHAP-based explainability visualizations. The deployment prioritizes usability, accessibility, and interpretability.

## IV. RESULTS AND DISCUSSION

### A. Model Performance Across Sampling Strategies

The performance of six classifiers KNN, Logistic Regression, Random Forest, XGBoost, Neural Network, and LightGBM was evaluated under three class balancing scenarios: original dataset, SMOTE oversampling, and random undersampling. Figures 1-3 summarize the results for each strategy using five evaluation metrics: accuracy, precision, recall, F1-score, and ROC-AUC.

As shown in Figure 1, performance on the original dataset favored accuracy but suffered in recall, indicating the models failed to detect many diabetic cases. SMOTE improved recall significantly (Figure 2), but led to reduced precision and accuracy. Undersampling (Figure 3) achieved the best recall across all models, especially with Logistic Regression, LightGBM, and XGBoost, making it the preferred approach for minimizing false negatives.

| Model | Accuracy | Precision | Recall | F1 | ROC-AUC |
|---|---|---|---|---|---|
| KNN | 0.8298 | 0.4142 | 0.2214 | 0.2886 | 0.7077 |
| LightGBM | 0.8543 | 0.6086 | 0.1818 | 0.2811 | 0.8235 |
| XGBoost | 0.8511 | 0.5693 | 0.1849 | 0.2792 | 0.8192 |
| Random Forest | 0.8413 | 0.4776 | 0.1943 | 0.2763 | 0.7793 |
| Logistic Regression | 0.8493 | 0.5544 | 0.1677 | 0.2575 | 0.8141 |
| Neural Network | 0.8524 | 0.5994 | 0.1603 | 0.2529 | 0.8177 |

Figure 1. Final Model Performance on Original Dataset

Final Model Comparison Table (With SMOTE):

| Model | Accuracy | Precision | Recall | F1 | ROC-AUC |
|---|---|---|---|---|---|
| Neural Network | 0.7807 | 0.3614 | 0.5304 | 0.4299 | 0.7798 |
| Logistic Regression | 0.7501 | 0.3332 | 0.6025 | 0.4291 | 0.7726 |
| LightGBM | 0.8224 | 0.4293 | 0.4236 | 0.4264 | 0.7997 |
| XGBoost | 0.8209 | 0.4254 | 0.4240 | 0.4247 | 0.7938 |
| Random Forest | 0.8056 | 0.3792 | 0.3885 | 0.3838 | 0.7601 |
| KNN | 0.7092 | 0.2768 | 0.5367 | 0.3652 | 0.7013 |

Figure 2. Final Model Performance with SMOTE Oversampling

Final Model Comparison Table (With Undersampling):

| Model | Accuracy | Precision | Recall | F1 | ROC-AUC |
|---|---|---|---|---|---|
| Logistic Regression | 0.7275 | 0.3351 | 0.7599 | 0.4651 | 0.8143 |
| LightGBM | 0.7164 | 0.3283 | 0.7834 | 0.4627 | 0.8222 |
| XGBoost | 0.7143 | 0.3258 | 0.7792 | 0.4595 | 0.8157 |
| Neural Network | 0.7249 | 0.3297 | 0.7408 | 0.4563 | 0.8080 |
| Random Forest | 0.7048 | 0.3137 | 0.7522 | 0.4427 | 0.7942 |
| KNN | 0.6791 | 0.2885 | 0.7219 | 0.4122 | 0.7517 |

Figure 3. Final Model Performance with Undersampling

### B. Statistical Validation of Model Differences

To verify the significance of performance differences between models across sampling strategies, a one-way ANOVA test was conducted on recall scores obtained from stratified cross-validation. For both SMOTE and random undersampling, the resulting p-values were well below the 0.05 threshold, indicating that the choice of model significantly influenced recall outcomes.

Specifically, for SMOTE, the ANOVA test yielded a p-value of $2.13 \times 10^{-24}$, and for random undersampling, the p-value was $5.39 \times 10^{-10}$, confirming that the observed recall variations are statistically significant and not due to random fluctuations. These findings validate that differences among model performances are reliable and merit further exploration.

To identify which specific model pairs differed significantly, a Tukey HSD (Honestly Significant Difference) post-hoc test was applied. This method controls the family-wise error rate and provides pairwise comparisons of model recall means. The results are visualized using compact summary tables that include mean differences, confidence intervals, and rejection flags indicating statistical significance (True/False).

Visual inspection of recall scores under SMOTE and undersampling (Figures. 4 and 6) highlights noticeable differences between classifiers. While some models like Random Forest and KNN perform reasonably well in isolated conditions, their variability and sensitivity to sampling techniques reduce reliability. In contrast, LightGBM consistently achieved high recall scores, particularly under undersampling, and displayed minimal variance across folds.

Under both sampling conditions, KNN, Logistic Regression, and Random Forest exhibited the most pronounced performance gaps compared to LightGBM, XGBoost, and Neural Network models. The Tukey HSD results (Figures 5 and 7) confirm these differences with statistically significant rejection flags in the majority of pairwise comparisons involving underperforming models.

The prioritization of recall as the evaluation metric is critical in this health-focused context. False negatives i.e., predicting a non-diabetic status for a truly diabetic individual can result in missed early interventions and delayed care. Therefore, selecting a model that maximizes recall while maintaining interpretability and deployment feasibility is essential. These statistical findings reinforce the robustness of the chosen LightGBM model under undersampling, supporting its deployment in the final web application.

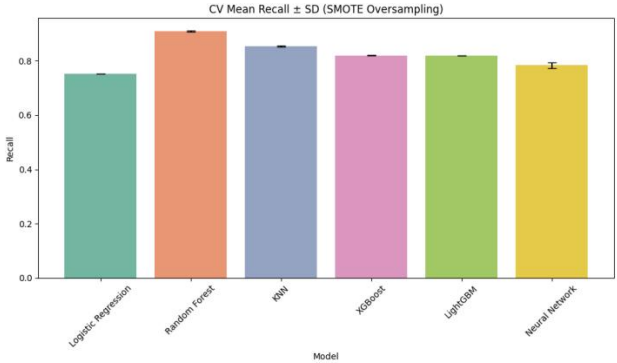

Figure 4. Cross-Validation Recall Scores with SMOTE (Mean ± SD)

```
                Multiple Comparison of Means - Tukey HSD, FWER=0.05
========================================================================================
group1              group2              meandiff  p-adj   lower    upper   reject
----------------------------------------------------------------------------------------
```

| group1 | group2 | meandiff | p-adj | lower | upper | reject |
|---|---|---|---|---|---|---|
| KNN | LightGBM | -0.0346 | 0.0 | -0.0443 | -0.0249 | True |
| KNN | Logistic Regression | -0.1017 | 0.0 | -0.114 | -0.092 | True |
| KNN | Neural Network | -0.0694 | 0.0 | -0.0791 | -0.0596 | True |
| KNN | Random Forest | 0.0553 | 0.0 | 0.0456 | 0.065 | True |
| KNN | XGBoost | -0.0342 | 0.0 | -0.0439 | -0.0245 | True |
| LightGBM | Logistic Regression | -0.0671 | 0.0 | -0.0768 | -0.0574 | True |
| LightGBM | Neural Network | -0.0348 | 0.0 | -0.0445 | -0.025 | True |
| LightGBM | Random Forest | 0.0899 | 0.0 | 0.0802 | 0.0996 | True |
| LightGBM | XGBoost | 0.0004 | 1.0 | -0.0093 | 0.0101 | False |
| Logistic Regression | Neural Network | 0.0323 | 0.0 | 0.0226 | 0.042 | True |
| Logistic Regression | Random Forest | 0.157 | 0.0 | 0.1473 | 0.1667 | True |
| Logistic Regression | XGBoost | 0.0675 | 0.0 | 0.0578 | 0.0772 | True |
| Neural Network | Random Forest | 0.1247 | 0.0 | 0.1149 | 0.1344 | True |
| Neural Network | XGBoost | 0.0352 | 0.0 | 0.0254 | 0.0449 | True |
| Random Forest | XGBoost | -0.0895 | 0.0 | -0.0992 | -0.0798 | True |

Figure 5. Tukey HSD Post-Hoc Comparison of Model Recall Scores (SMOTE)

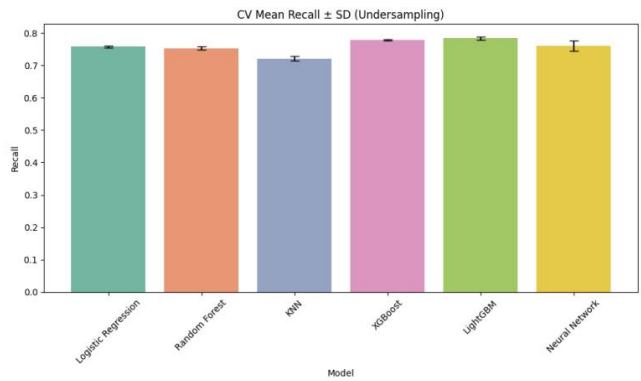

Figure 6. Cross-Validation Recall Scores with Undersampling (Mean ± SD)

```
                Multiple Comparison of Means - Tukey HSD, FWER=0.05
========================================================================================
group1              group2              meandiff  p-adj   lower    upper   reject
----------------------------------------------------------------------------------------
```

| group1 | group2 | meandiff | p-adj | lower | upper | reject |
|---|---|---|---|---|---|---|
| KNN | LightGBM | 0.0623 | 0.0 | 0.0455 | 0.0792 | True |
| KNN | Logistic Regression | 0.0362 | 0.0 | 0.0193 | 0.053 | True |
| KNN | Neural Network | 0.0392 | 0.0 | 0.0223 | 0.0561 | True |
| KNN | Random Forest | 0.0315 | 0.0001 | 0.0146 | 0.0483 | True |
| KNN | XGBoost | 0.0573 | 0.0 | 0.0404 | 0.0742 | True |
| LightGBM | Logistic Regression | -0.0262 | 0.0009 | -0.0431 | -0.0093 | True |
| LightGBM | Neural Network | -0.0232 | 0.0034 | -0.04 | -0.0063 | True |
| LightGBM | Random Forest | -0.0309 | 0.0001 | -0.0478 | -0.014 | True |
| LightGBM | XGBoost | -0.0051 | 0.9351 | -0.022 | 0.0118 | False |
| Logistic Regression | Neural Network | 0.003 | 0.9931 | -0.0139 | 0.0199 | False |
| Logistic Regression | Random Forest | -0.0047 | 0.9522 | -0.0216 | 0.0122 | False |
| Logistic Regression | XGBoost | 0.0211 | 0.0085 | 0.0042 | 0.038 | True |
| Neural Network | Random Forest | -0.0077 | 0.7186 | -0.0246 | 0.0092 | False |
| Neural Network | XGBoost | 0.0181 | 0.0307 | 0.0012 | 0.035 | True |
| Random Forest | XGBoost | 0.0258 | 0.001 | 0.0089 | 0.0427 | True |

Figure 7. Tukey HSD Post-Hoc Comparison of Model Recall Scores (Undersampling)

### C. SHAP-Based Global and Local Interpretability

To ensure transparency in model predictions and gain insights into feature contributions, SHAP (SHapley Additive exPlanations) was employed. SHAP is based on cooperative game theory and attributes model output to each feature by calculating their marginal contribution across all possible feature combinations. It enables both **global** (dataset-level) and **local** (individual-level) interpretability.

#### 1) Global Feature Importance

A SHAP summary plot was generated to identify which features most influenced diabetes predictions across the dataset. As shown in Figure 8, the most impactful features were:

- **Risk Factor Count:** An engineered variable aggregating chronic conditions like HighBP, HighChol, Stroke, and HeartDiseaseorAttack.
- **Physical Health (PhysHlth):** Denotes the number of days with poor physical health.
- **HighBP and General Health (GenHlth):** Indicate chronic disease presence and perceived well-being.
- **Difficulty Walking (DiffWalk) and Cholesterol Check (CholCheck):** Reflect mobility and engagement with preventive care.

The SHAP summary plot ranks features based on their mean absolute SHAP values, reflecting their overall influence on model predictions. Features positioned higher in the plot contribute more significantly to the prediction outcome. The horizontal spread of each feature indicates whether its impact pushes the prediction toward the diabetic (positive SHAP value) or non-diabetic class (negative SHAP value).

For example, individuals with high Risk Factor Count or poor physical health (PhysHlth) typically exhibit strong positive contributions toward diabetes classification. The color gradient encodes the actual feature values, with red representing higher values and blue indicating lower ones. This visualization provides a comprehensive and interpretable overview of model behavior, reinforcing the relevance of selected predictors in a clinically meaningful way.

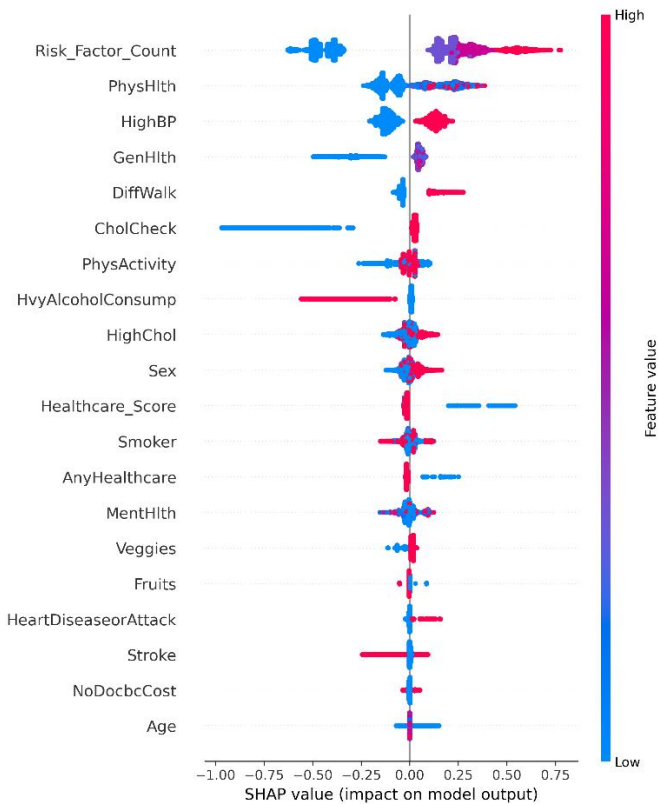

Figure 8. SHAP Summary Plot of Global Feature Importance

**2) Local Explanation with SHAP Waterfall Plot**

Figure 9 displays a SHAP waterfall plot that explains how individual features influenced the model's prediction for a specific case. The base value (1.393), which represents the average model output in log-odds, is adjusted by the SHAP values of each feature to reach the final model prediction of 2.27 log-odds for this individual.

Each numeric value next to a feature (e.g., +0.24 or –0.03) represents the feature's additive contribution to the prediction. A positive SHAP value indicates that the feature increased the predicted risk, while a negative value decreased it. For example, high values for PhysHlth (15 days), Risk Factor Count (2), and the presence of HighBP (1) and DiffWalk (1) had the strongest positive impact, pushing the prediction well above the average. In contrast, Smoker (1) showed a mild negative contribution, slightly lowering the output.

The final log-odds score is internally converted to a probability for classification, but this intermediate representation helps to understand which features had the greatest effect. Such local explanations not only improve transparency but also support clinically relevant, personalized interpretation—highlighting risk factors specific to an individual case.

This type of local explanation is particularly valuable in healthcare applications, where understanding why a prediction was made is as important as the prediction itself. By quantifying the contribution of each feature, SHAP enables clinicians, public health professionals, or end-users to trace the reasoning behind a high-risk or low-risk classification. In this case, the prediction can be clearly attributed to measurable health conditions such as elevated physical health concerns and comorbid risks. These insights can guide individualized interventions, promote informed decision-making, and build trust in machine learning systems deployed in sensitive domains like health risk assessment.

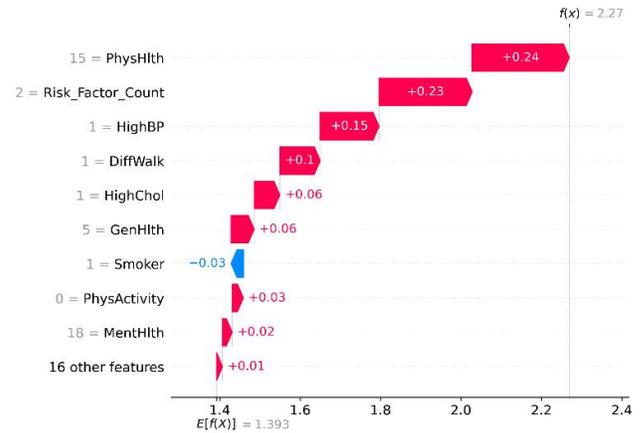

Figure 9. SHAP Waterfall Plot for an Individual Prediction

**D. LIME-Based Local Explanations**

To enhance model interpretability at the individual level, LIME (Local Interpretable Model-Agnostic Explanations) was applied. LIME generates explanations by approximating the model locally around a given prediction using a simple interpretable model (typically linear regression). This allows for clear visualization of which features contributed most to a specific classification.

In this study, a representative prediction was selected for analysis. As shown in **Figure 10**, the model predicted a **92% probability of diabetes** for the chosen individual. The bar chart highlights features that contributed to this prediction:

- **BMI > 32**, **GenHlth > 3**, and **Risk Factor Count = 2** were the **strongest contributors** driving the prediction toward the diabetic class.

- Other contributing factors included **high blood pressure**, **difficulty walking**, and **elevated cholesterol**, all of which increased the likelihood of a positive classification.

- Features such as **income ≤ 5**, **moderate physical health issues**, and **age between 8 and 10 (coded bin)** also reinforced the diabetic prediction.

On the right side of the figure, the corresponding feature values used by the model are displayed, confirming the individual had elevated values for **BMI (40.0)**, **GenHlth (5)**, and multiple chronic conditions.

This visual explanation not only confirms the decision rationale but also supports user trust and transparency in a healthcare context. By clearly identifying the top risk-driving factors, LIME empowers patients and clinicians to focus on the most influential health areas.

In healthcare applications, the ability to generate such individualized explanations is essential for fostering transparency and patient engagement. LIME's local approximations allow users to explore how specific health factors influence model output in a manner that aligns with clinical reasoning. This makes the tool particularly effective for patient education, shared decision-making, and early intervention planning. By surfacing interpretable, instance-level justifications, LIME enables both clinicians and patients to move beyond black-box predictions toward actionable, personalized healthcare insights.

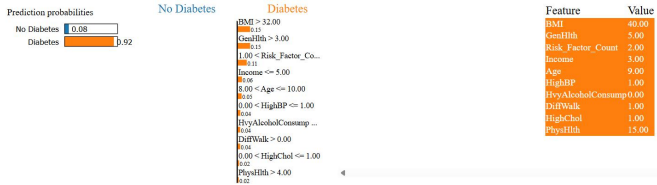

Figure 10. LIME Explanation for an Individual Prediction

### E. Comorbidity Insights

Understanding how diabetes interacts with other chronic health conditions is critical for assessing overall patient risk. To explore these associations, a Pearson correlation analysis was conducted between the diabetes outcome variable and other risk factors such as High Blood Pressure, High Cholesterol, Stroke, and Heart Disease.

As illustrated in Figure 11, the strongest comorbidity was observed between diabetes and high blood pressure ($r = 0.26$), followed by high cholesterol ($r = 0.20$), heart disease or heart attack ($r = 0.17$), and stroke ($r = 0.10$). These moderate but meaningful correlations align with established clinical research, which identifies these conditions as common complications or co-existing risks in diabetic individuals.

This comorbidity analysis not only reinforces the predictive importance of these variables in the model but also enhances the real-world relevance of the deployed tool. In the application, users flagged as diabetic can be simultaneously alerted to increased risks of cardiovascular disease, thereby promoting preventive awareness and early clinical engagement.

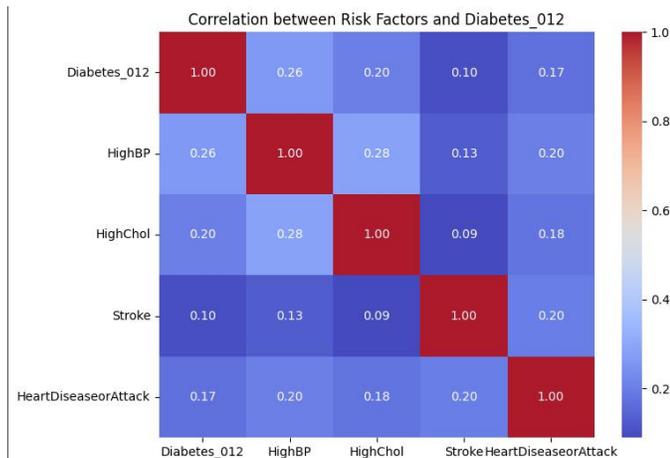

Figure 11. Correlation Heatmap Between Diabetes and Related Chronic Conditions

### D. Interactive Web Interface and User Experience

To make the predictive model accessible and user-friendly, an interactive web application was developed using the Dash framework. The application features a multi-step form that collects demographic, lifestyle, medical history, and healthcare-related information to generate personalized diabetes risk predictions. The interface was designed with **glassmorphism styling** for modern aesthetics and clarity.

Users are guided through the following sequential steps:

- **Step 1: Basic Information** (Age, Sex, BMI – Figure 12)
- **Step 2: Lifestyle Habits** (Smoking, alcohol consumption, fruit/vegetable intake, physical activity – Figure 13)
- **Step 3: Medical History** (High BP, High Chol, Stroke, Heart Disease – Figure 14)
- **Step 4: Healthcare Access & Physical Condition** (General health, mental/physical health days, walking difficulty – Figure 15)
- **Step 5: Socioeconomic Details** (Education and Income levels – Figure 16)

After all inputs are submitted, users are shown the **prediction result**, **lifestyle and healthcare scores**, and **explainable insights** derived from SHAP and LIME (Figure 18–21). The final screen also provides:

- **Top SHAP features** influencing the prediction
- **LIME interpretation** for the specific individual
- **Personalized health improvement suggestions**
- **Comorbidity risks and recommendations**

This user-friendly interface bridges technical output with public usability, enabling non-technical users to understand their diabetes risk and contributing factors.

The platform emphasizes accessibility and interpretability, combining modern UI design with explainable AI outputs. Visual elements such as SHAP and LIME plots, lifestyle scoring, and comorbidity insights help users understand both their risk and contributing factors. This integration of prediction and explanation supports informed decision-making in both clinical and public health contexts.

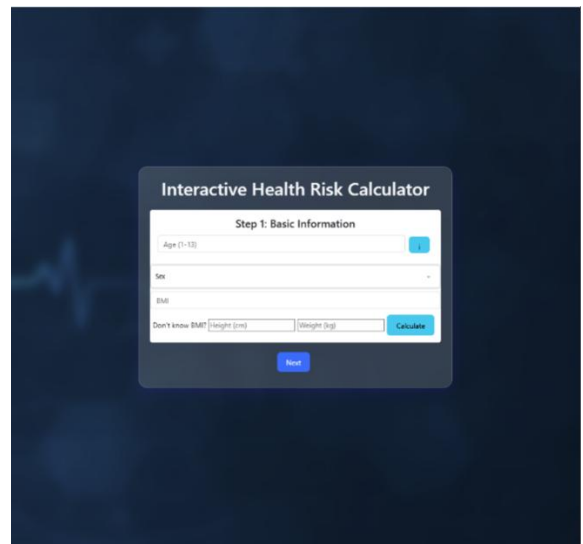

Figure 12 . Step 1: Basic Information Input

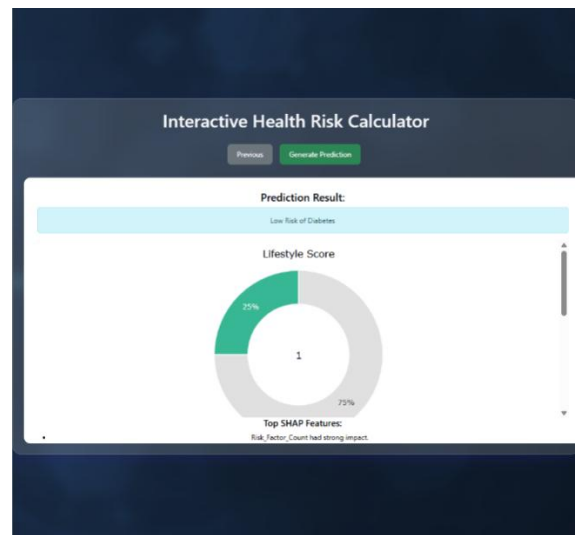

Figure 13. Step 2: Lifestyle Habits Form

Figure 16. Step 5: Education & Income Level Form

Figure 14. Step 3: Medical History Form

Figure 17. Prediction Submission Interface

Figure 15. Step 4: Healthcare & Physical
Condition Form

Figure 18. Prediction Result with Lifestyle Score

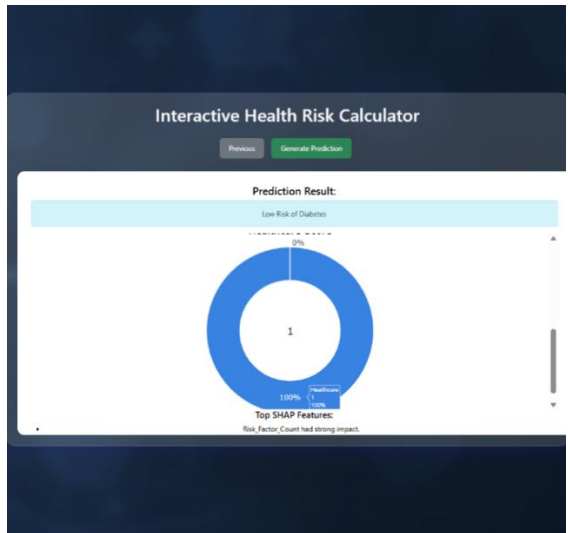

Figure 19. Prediction Result with Healthcare Score

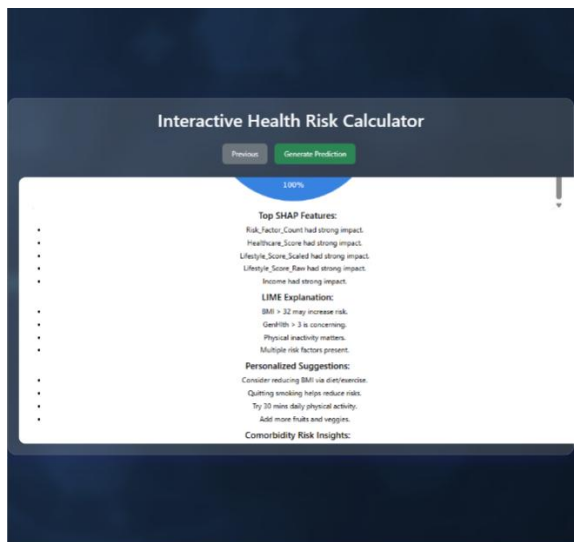

Figure 20. Top SHAP Features and LIME Explanations

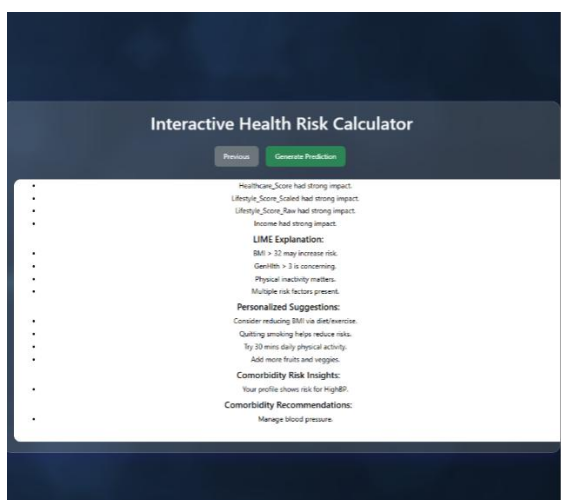

Figure 21. Personalized Suggestions and
Comorbidity Insights

## V. Conclusion and Future Work

This study developed an explainable, user-centered diabetes risk prediction system using the 2015 BRFSS dataset and machine learning. The final LightGBM model, selected through extensive cross-validation and statistical testing, demonstrated superior recall performance under an undersampling strategy. To promote model transparency, SHAP and LIME were employed to provide both global and local explanations, offering insights into the key drivers of diabetes predictions at both the population and individual levels.

A key contribution of this work is the deployment of a web-based Health Risk Calculator that transforms static predictions into actionable health insights. The multi-step interface not only guides users through demographic, lifestyle, and clinical input collection but also delivers tailored predictions, interpretability visuals, and comorbidity alerts. This integration of predictive analytics and explainable AI makes the system practical and accessible for real-world use.

Future Work will focus on expanding the model's scope beyond diabetes to include multi-disease risk prediction, using longitudinal or more recent datasets for improved temporal relevance. Integration with electronic health records (EHRs), inclusion of real-time wearable data, and implementation of multi-language support are also planned to improve usability and accuracy. Additionally, incorporating user feedback loops and external clinical validation will further strengthen the model's reliability and applicability in healthcare settings.

The explainable nature of the system ensures that predictions are not only accurate but also interpretable, helping bridge the gap between black-box machine learning models and clinical trust. By surfacing clear reasoning for each prediction, the system empowers both patients and healthcare providers to make informed decisions. This transparency is critical for fostering the adoption of AI-driven tools in healthcare environments, particularly when applied to chronic disease prevention and early detection.

Moreover, this research contributes to the growing field of interpretable machine learning by demonstrating a complete pipeline from data preprocessing and model selection to post hoc explanation and real-time deployment. The modular design of the framework allows for easy adaptation to other health conditions and datasets. As health data becomes more complex and high-dimensional, the need for interpretable, interactive, and scalable prediction systems will only increase, positioning this work as a strong foundation for future advances in patient-centered predictive analytics.

In summary, this study not only highlights the predictive capabilities of modern machine learning techniques but also emphasizes the importance of usability, interpretability, and personalization in digital health applications. By integrating statistical rigor with user-focused design and explainable AI, the system demonstrates a balanced approach to technology-driven health risk assessment. As AI continues to shape the future of healthcare, solutions like this provide a template for building transparent, inclusive, and impactful tools that align with both clinical standards and public health needs.

## VI. ACKNOWLEDGEMENT

The author would like to sincerely thank **Dr. Mariofanna Milanova**, Professor of Computer Science at the University of Arkansas at Little Rock and IEEE Senior Member, for her invaluable mentorship and technical guidance throughout the development of this research project. The author also wishes to thank **Dr. Francesco Cavarretta**, Assistant Professor in the Department of Computer Science and Research Fellow at the Emerging Analytics Center at the University of Arkansas at Little Rock, for his insightful feedback and support. Their expertise in artificial intelligence, machine learning, and computational modeling played a crucial role in shaping both the predictive framework and the interpretability components of the system. This work was conducted as part of a graduate research project in the Department of Computer and Information Sciences at the University of Arkansas at Little Rock.

## REFERENCES

[1] M. M. Chowdhury, R. S. Ayon, and M. S. Hossain, "Diabetes diagnosis through machine learning: Investigating algorithms and data augmentation for class imbalanced BRFSS dataset," Preprint, Dept. of Mathematics and Statistics, Texas Tech Univ., Lubbock, TX, USA, and Dept. of Electronics and Telecommunication Eng., Rajshahi Univ. of Engineering and Technology, Rajshahi, Bangladesh, 2023.

[2] Z. Liu, Q. Zhang, H. Zheng, S. Chen, and Y. Gong, "A comparative study of machine learning approaches for diabetes risk prediction: Insights from SHAP and feature importance," Preprints, Nov. 19, 2024. [Online]. Available: https://doi.org/10.20944/preprints202411.1265.v1

[3] S. Ahmed, M. S. Kaiser, M. S. Hossain, and K. Andersson, "A comparative analysis of LIME and SHAP interpreters with explainable ML-based diabetes predictions," IEEE Access, vol. 12, pp. 87912–87925, 2025. doi: 10.1109/ACCESS.2024.3422319.

[4] B. Nguyen and Y. Zhang, "A comparative study of diabetes prediction based on lifestyle factors using machine learning," unpublished manuscript, 2025.

[5] H. Nguyen, H. Cao, V. Nguyen, and D. Pham, "Evaluation of Explainable Artificial Intelligence: SHAP, LIME, and CAM," *ResearchGate preprint*, 2021. [Online]. Available: https://www.researchgate.net/publication/362165633

[6] A. L. R. Agahan, M. S. A. Magboo, and V. P. C. Magboo, "Predicting the Risk of Diabetes Using Explainable Artificial Intelligence," in *Proc. 2023 Int. Conf. on Electrical, Computer and Energy Technologies (ICECET)*, 2023, pp. 1–6. doi: 10.1109/ICECET58911.2023.10389419.

[7] N. Gandhi and S. Mishra, "Explainable AI for healthcare: A study for interpreting diabetes prediction," in Proc. Int. Conf. on Machine Learning and Big Data Analytics (ICMLBDA), Cham, Switzerland: Springer, 2022, pp. 95–105.

[8] Y. Ramon, D. Martens, F. Provost, and T. Evgeniou, "A comparison of instance-level counterfactual explanation algorithms for behavioral and textual data: SEDC, LIME-C and SHAP-C," Advances in Data Analysis and Classification, vol. 14, no. 4, pp. 801–819, Dec. 2020.

[9] Y. Zhao, J. K. Chaw, M. C. Ang, M. M. Daud, and L. Liu, "A diabetes prediction model with visualized explainable artificial intelligence (XAI) technology," in *Advances in Visual Informatics (Lecture Notes in Computer Science)*, vol. 14322, H. B. Zaman, Ed., Singapore: Springer, 2024. doi: 10.1007/978-981-99-7339-2_52.

[10] A. Priyadarshini and J. Aravinth, "Correlation based breast cancer detection using machine learning," in Proc. Int. Conf. on Recent Trends in Electronics, Information & Communication Technology (RTEICT), 2021, pp. 499–504.

[11] A. Yahyaoui, A. Jamil, J. Rasheed, and M. Yesiltepe, "A decision support system for diabetes prediction using machine learning and deep learning techniques," in Proc. 1st Int. Informat. Softw. Eng. Conf. (UBMYK), 2019, pp. 1–4.

[12] A. Mujumdar and V. Vaidehi, "Diabetes prediction using machine learning algorithms," Procedia Computer Science, vol. 165, pp. 292–299, 2019.

[13] N. Fazakis, O. Kocsis, E. Dritsas, S. Alexiou, N. Fakotakis, and K. Moustakas, "Machine learning tools for long-term type 2 diabetes risk prediction," IEEE Access, vol. 9, pp. 103737–103757, 2021.

[14] M. A. Sarwar, N. Kamal, W. Hamid, and M. A. Shah, "Prediction of diabetes using machine learning algorithms in healthcare," in Proc. 24th Int. Conf. on Automation and Computing (ICAC), Sep. 2018, pp. 1–6.

[15] M. U. Emon, M. S. Keya, M. S. Kaiser, M. A. Islam, T. Tanha, and M. S. Zulfiker, "Primary stage of diabetes prediction using machine learning approaches," in Proc. Int. Conf. on Artificial Intelligence and Smart Systems (ICAIS), Mar. 2021, pp. 364–367.

[16] Z. Q. Lin, M. J. Shafiee, S. Bochkarev, M. St. Jules, X. Y. Wang, and A. Wong, "Do explanations reflect decisions? A machine-centric strategy to quantify the performance of explainability algorithms," arXiv preprint arXiv:1910.07387, 2019.

[17] J. J. Bigna and J. J. Noubiap, "The rising burden of non-communicable diseases in sub-Saharan Africa," The Lancet Global Health, vol. 7, no. 10, pp. e1295–e1296, 2019.

[18] L. Breiman, "Random forests," Machine Learning, vol. 45, pp. 5–32, 2001.

[19] T. A. Buchanan, A. H. Xiang, et al., "Gestational diabetes mellitus," *The Journal of Clinical Investigation*, vol. 115, no. 3, pp. 485–491, 2005.

[20] A. Budreviciute, S. Damiati, D. K. Sabir, K. Onder, P. Schuller-Goetzburg, G. Plakys, A. Katileviciute, S. Khoja, and R. Kodzius, "Management and prevention strategies for non-communicable diseases (NCDs) and their risk factors," *Frontiers in Public Health*, vol. 8, p. 788, 2020.

[21] J. Burez and D. Van den Poel, "Handling class imbalance in customer churn prediction," Expert Systems with Applications, vol. 36, no. 3, pp. 4626–4636, 2009.

[22] C. J. Caspersen, G. D. Thomas, L. A. Boseman, G. L. Beckles, and A. L. Albright, "Aging, diabetes, and the public health system in the United States," *American Journal of Public Health*, vol. 102, no. 8, pp. 1482–1497, 2012.